# Extreme Speech Classification in the Era of LLMs: Exploring Open-Source and Proprietary Models

Sarthak Mahajan         Nimmi Rangaswamy

sarthak.mahajan@research.iiit.ac.in  nimmi.rangaswamy@iiit.ac.in

**Abstract.** In recent years, widespread internet adoption and the growth in userbase of various social media platforms have led to an increase in the proliferation of extreme speech online. While traditional language models have demonstrated proficiency in distinguishing between neutral text and non-neutral text (i.e. extreme speech), categorizing the diverse types of extreme speech presents significant challenges [1][2]. The task of extreme speech classification is particularly nuanced, as it requires a deep understanding of socio-cultural contexts to accurately interpret the intent of the language used by the speaker. Even human annotators often disagree on the appropriate classification of such content, emphasizing the complex and subjective nature of this task [3]. The use of human moderators also presents a scaling issue, necessitating the need for automated systems for extreme speech classification. The recent launch of ChatGPT has drawn global attention to the potential applications of Large Language Models (LLMs) across a diverse variety of tasks. Trained on vast and diverse corpora, and demonstrating the ability to effectively capture and encode contextual information, LLMs emerge as highly promising tools for tackling this specific task of extreme speech classification. In this paper, we leverage the Indian subset of the extreme speech dataset from [3] to develop an effective classification framework using LLMs. We evaluate open-source Llama models against closed-source OpenAI models, finding that while pre-trained LLMs show moderate efficacy, fine-tuning with domain-specific data significantly enhances performance, highlighting their adaptability to linguistic and contextual nuances. Although GPT-based models outperform Llama models in zero-shot settings, the performance gap disappears after fine-tuning.

**Keywords:** Extreme Speech, Large Language Models, Llama, Open Source.

## 1   Introduction

With increasing polarization of society along the lines of religion, race, gender, ethnicity, etc., there has been a corresponding increase in hate speech online. Its manifestations vary across countries, reflecting their unique sociopolitical and cultural contexts- racial and sexual orientation-based in the United States, anti-immigrant in Germany, and religious or caste-based in India.

Social media companies have invested heavily in content moderation systems to address hate speech, yet the scale of user-generated content renders manual moderation insufficient, making automated systems essential for scalable and effective detec-tion [5][6]. While prior research has focused on dataset creation [3], NLP-based detection [8][9], and target identification [10], accurate hate speech classification remains challenging, requiring nuanced understanding of sociocultural contexts.

### 1.1 Contributions

This study focuses on the Indian subset of the Xtreme Speech Dataset, introduced by Maronikolakis et al. [3], to evaluate the performance of large language models (LLMs) in classifying extreme speech into three categories: derogatory extreme speech, exclusionary extreme speech, and dangerous speech. Utilizing both open-source and proprietary closed source LLMs, we compare multiple approaches, including zero-shot classification, fine-tuned models, fine-tuning with Direct Preference Optimization (DPO), and an ensemble of fine-tuned models. We also explore whether open-source models (specifically the Llama family of models) are a viable alternative to their more famous closed source counterparts like the GPT models by OpenAI.

### 1.2 Defining Extreme Speech

Hate speech, broadly defined as "public speech that expresses hate or encourages violence towards a person or group based on some characteristic such as race, color, religion, national origin, sexual orientation, disability, or other traits", is a concerning issue on digital platforms [11]. Udupa et al. [12] take an anthropological perspective for defining extreme speech as speech that pushes the boundaries of civil language. While hate speech is often framed in legal and political terms, extreme speech encompasses a wider range of harmful, provocative, or exclusionary discourse that may not meet legal definitions of hate [3].

In this paper, we adopt the definition of extreme speech as proposed by Pohjonen and Udupa, who define it as "speech acts that push the boundaries of acceptable speech by expressing views that challenge or threaten core values, identities and social arrangements." [12] This definition allows us to capture a broader range of problematic speech beyond just hate speech, including speech that may not necessarily be hateful but is still harmful, exclusionary or divisive.

**Types of Extreme Speech.** Maronikolakis et al. [3] categorize extreme speech into:

- Derogatory speech: Speech that can be considered uncivil and offensive to any community, person, group, or institution. This form of speech may also be part of a form of protest against prevailing social norms or political powers.
- Dangerous speech: Speech that has the potential to lead to physical harm or violence.
- Exclusionary speech: Speech that segregates or excludes a certain group or community. This exclusion is often promoted through the use of humour in order to normalize it.

### 1.3 LLMs for Extreme Speech Moderation

In recent years, the advent of powerful LLMs has revolutionized the field of natural language processing. These models, trained on huge and diverse datasets, have achieved state-of-the-art performance in areas such as text generation, question answering, and sentiment analysis [4] [13]. Their ability to generalize across domains through fine-tuning or prompt-based conditioning has made them indispensable tools for both academic research and industry applications [14]. LLMs have been employed to generate coherent and contextually relevant text, answer complex queries with high accuracy, and analyze sentiment in social media data with remarkable precision [15].

**Open source versus Closed source LLMs.** We focus specifically on open-source models due to their transparency, which provides deeper insights into model behavior and limitations compared to proprietary, black-box alternatives. For extreme speech moderation, model choice is critical not only for performance but also for reproducibility and ethical considerations. Open-source models, such as Meta AI's Llama family, offer full access to their codebase, weights, and training methodologies, enabling reproducibility and facilitating scientific rigor [7].

## 2 Related Work

Earlier works in hate speech focused on detecting hate speech on social media platforms [8][20] [21]. Some of these works have also contributed to data collection and annotation [20][21] to assist in hate speech detection.

Others have expanded on the data collection and curation efforts, by contributing multi-lingual datasets [22][23][24]. In conjunction, researchers have looked at hate speech detection in multilingual settings, and examined transfer learning techniques to improve performance on low-resource languages [25].

Complementing these efforts on multilingual dataset curation, a recent work by Maronikolakis et al. introduced the Xtreme Speech Dataset, which expands the scope beyond just hate speech to include a broader range of problematic speech, called extreme speech, such as derogatory extreme speech, exclusionary extreme speech and dangerous speech [3].

Researchers have also looked at unique challenges posed by hate speech moderation, focusing on aspects like the ambiguous and context-dependent nature of hate speech, cultural nuances of language, and the dynamic nature of hate speech [3][26].

Various machine learning techniques have been explored for detecting hate speech online, using approaches ranging from traditional ML models like SVM, Naive Bayes [27] to more recent deep learning methods [4] [8].

In recent years, with the growing popularity of LLMs in various NLP tasks, researchers have explored using LLMs for text classification [28]. There is a growing use of LLMs in content moderation [29], owing to their ability to generalize across tasks, having been pretrained on vast and diverse datasets.

Our work focuses on the Indian part of the dataset introduced by Maronikolakis et al., and shows how LLMs, trained on a massive corpus of data, have the nuance and contextual understanding to discriminate between these fine-grained categories of extreme speech. We look at the performance of pre-trained LLMs through zero shot inference, and contrast it with performance jumps achieved by fine tuning the model by giving it examples of extreme speech in the Indian context.

## 3 Dataset

We use the Indian partition of the Xtreme Speech Dataset [3], which includes 4,933 samples (after removing duplicates from the original dataset) across three categories. The dataset is split into training (64%), DPO/ensemble (16%), and testing (20%) sets, with representative sampling to maintain consistent class distributions across splits.

Table 1. Distribution of dataset by label and across train, ensemble/DPO, test sets

| Label | Train | Ensemble/DPO | Test | Full dataset |
|---|---|---|---|---|
| Derogatory extreme speech | 1438 | 341 | 411 | 2190 |
| Exclusionary extreme speech | 904 | 214 | 279 | 1397 |
| Dangerous speech | 814 | 235 | 297 | 1346 |
| Total | 3156 | 790 | 987 | 4933 |

## 4 Methodology

We evaluate the Llama family of models, which are pretrained on diverse multilingual corpora and have demonstrated strong performance across various NLP tasks. For benchmarking, we also include GPT-4o and GPT-4o-mini to compare closed-source performance with open-source alternatives. To align the loss function with classification accuracy, we encode class labels as 0, 1, and 2 for derogatory, exclusionary, and dangerous speech, respectively[1].

### 4.1 Inference through Zero shot

We evaluate the following models: Open Source - Llama 3.1 8B, Llama 3.2 1B, Llama 3.2 3B, Llama 3.3 70B, and Closed Source - GPT 4o, and GPT 4o-mini. We used 4-bit quantized versions of the Llama models, reducing memory requirements and enabling faster inference. The models are prompted to categorize text from the test dataset into one of three predefined classes.

We tested two approaches: (1) predicting the class label (0/1/2) directly, and (2) first generating a justification followed by the predicted label. The latter approach improved classification performance, consistent with findings on "Chain-of-Thought" prompting, which enhances reasoning by breaking complex problems into intermediate steps [30]. Table 2 shows the results for zero-shot inference (with justifications).

---

[1] Code available at https://github.com/sarthak-mahajan/extreme_speech_classification

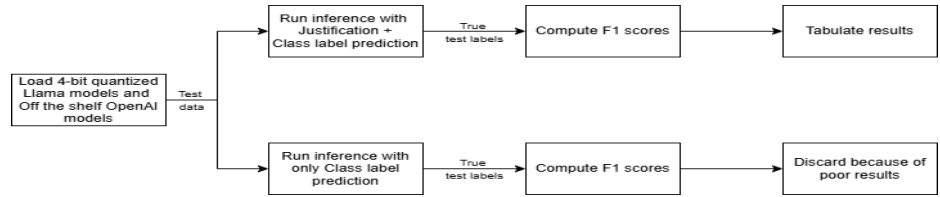

**Fig. 1.** Zero-Shot Inference

### 4.2 Inference through Supervised Fine-tuned (SFT) Models

We fine-tuned the models using training samples of extreme speech text paired with their corresponding labels. Two approaches were evaluated: one incorporating both justifications and labels during SFT, and another using only labels. Interestingly, the variant with justifications underperformed compared to the label-only approach. While justifications enhance zero-shot reasoning, they may not improve supervised learning, as the loss function does not explicitly optimize for classification accuracy when justifications are included. In contrast, training with only labels directly minimizes classification error, as the output is constrained to the label itself.

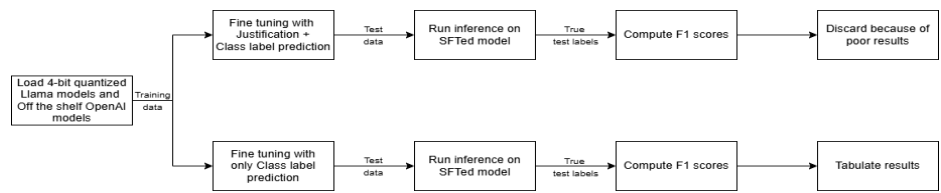

**Fig. 2.** Supervised Fine-tuned Models

### 4.3 Preference Optimization

We evaluated Direct Preference Optimization (DPO) [18] on the fine-tuned Llama 3.1 8B model to assess potential performance improvements. To construct the DPO dataset, we ran inference on the remaining 16% dataset using the SFT-ed Llama 3.1 8B model, extracting log probabilities for each output token (0/1/2, corresponding to class probabilities). For each example, the incorrect class with the highest probability was selected as the negative example, while the human annotated class label served as the positive example. DPO was then applied to the SFT-ed Llama 3.1 8B model using this dataset. The results of this experiment are shown in Table 3.

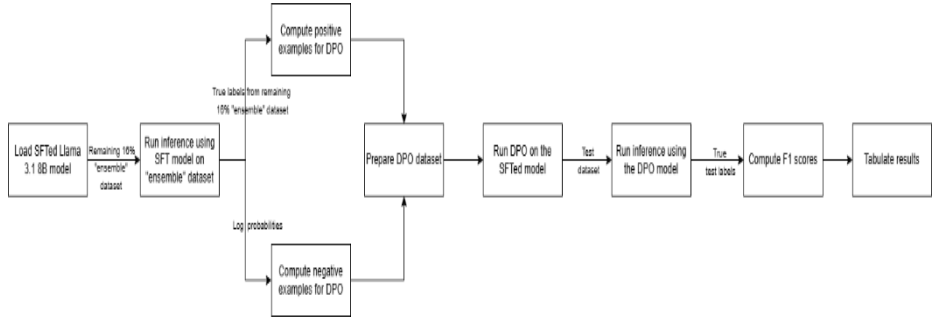

**Fig. 3.** Preference Optimization

### 4.4 Ensembling SFT-ed models

To further enhance performance, we explored ensembling the fine-tuned Llama models. We computed F1-macro scores for all four models on a held-out 16% dataset and tested two ensembling approaches: (1) weighting each model's predicted label by its F1-macro score and selecting the class with the highest weighted score, and (2) calculating a weighted average of class probabilities using F1-macro scores and selecting the class with the highest average probability. As shown in Table 3, both approaches yielded nearly identical results.

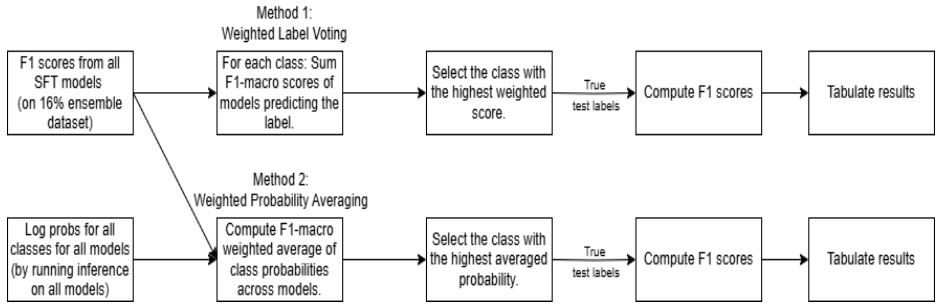

**Fig. 4.** Ensembling SFT models: Combining F1 macro & log probabilities to predict class label

## 5 Results and analysis

### 5.1 Zero Shot

LLMs demonstrate strong zero-shot performance, highlighting their ability to generalize across tasks without task-specific training [4]. Among Llama models, Llama 3.3 70B outperforms smaller variants, consistent with the established correlation between model size and performance. Larger models, with greater parameter capacity, capture complex linguistic patterns more effectively, leading to higher F1 scores.

Smaller Llama models perform comparably but at lower levels, likely due to their limited parameters. While 4-bit quantization improves accessibility by reducing memory requirements, it also constrains model capacity and precision [31], making it harder to capture nuanced linguistic and contextual variations in extreme speech.

GPT-4o achieves the highest performance, followed closely by GPT-4o-mini, excelling particularly in dangerous speech detection. This suggests that beyond model size, architectural advancements and pretraining strategies in GPT-4 contribute to its superior capabilities, especially in complex tasks like extreme speech classification.

**Table 2.** Zero shot – inference results on test set – F1 scores

| Label | Llama 3.2 1B | Llama 3.2 3B | Llama 3.1 8B | Llama 3.3 70B | gpt-4o-mini-2024-07-18 | gpt-4o-2024-08-06 |
|---|---|---|---|---|---|---|
| Derogatory | 45.28 | 59.72 | 63.01 | 64.51 | 67.72 | 71.45 |
| Exclusionary | 23.44 | 10.17 | 26.08 | 30.08 | 31.12 | 36.61 |
| Dangerous | 35.02 | 27.07 | 14.20 | 33.49 | 47.31 | 48.55 |
| F1-macro | 34.58 | 32.32 | 34.43 | 42.70 | 48.71 | 52.20 |

### 5.2 Supervised Fine-tuned Models

Fine-tuning significantly boosts performance across all models, demonstrating that incorporating cultural context helps models better identify extreme speech in the Indian dataset. Notably, fine-tuned Llama models, regardless of size (1B to 70B parameters), perform comparably, with even smaller models like Llama 3.1 1B and Llama 3.2 3B outperforming GPT-4o's zero-shot performance and nearing fine-tuned GPT-4o-mini. This underscores the importance of fine-tuning for domain-specific tasks.

### 5.3 Ensembling

As shown in Table 3, an ensemble of the fine-tuned Llama models does not outperform the individual fine-tuned models. We observed that the various fine-tuned models exhibit similar strengths and weaknesses; they perform well on the same classes and make comparable types of misclassifications. Ensembling typically relies on the principle of combining models with diverse error patterns to achieve superior performance [19]. However, in our case, the homogeneity in the fine-tuned models' performance limits the ensemble's ability to outperform individual models.

### 5.4 Preference Optimization

We observed no performance gains with DPO, indicating it offers no additional benefits over SFT for text classification, particularly for nuanced tasks like extreme speech. This aligns with DPO's design, which focuses on optimizing stylistic and preference-based text generation rather than discriminative tasks [17].

Table 3. Fine-tuned models, Ensemble with SFT-ed Llama models, DPO with Llama 3.1 8B - inference results on test set – F1 scores

| Label | Llama 3.2 1B | Llama 3.2 3B | Llama 3.1 8B | Llama 3.3 70B | gpt-4o-mini-2024-07-18 | Ensemble (F1 macro) | Ensemble (probability weighted F1 macro) | DPO (Llama 3.1 8B) |
|---|---|---|---|---|---|---|---|---|
| Derogatory | 71.18 | 72.14 | 74.85 | 72.86 | 75.65 | 75.38 | 74.97 | 75.02 |
| Exclusionary | 49.40 | 54.84 | 52.37 | 51.5 | 56.19 | 55.34 | 55.80 | 50.19 |
| Dangerous | 75.56 | 78.78 | 81.33 | 79.79 | 83.13 | 79.78 | 80.50 | 80.00 |
| F1-macro | 65.59 | 68.59 | 69.52 | 68.20 | 71.65 | 70.16 | 70.42 | 68.40 |

### 5.5 Benchmarking Against State-of-the-Art Classification Models

Table 4 compares the best F1 scores from Maronikolakis et al. [3] with those achieved by our fine-tuned models. Our results, including those from smaller Llama models, significantly outperform the state-of-the-art benchmarks, demonstrating the advantages of using LLMs for extreme speech classification. Notably, our results are more balanced across all classes, addressing the poor performance of langBERT and mBERT on exclusionary extreme speech.

Table 4. Best F1 scores from Maronikolakis et al [3] vs our fine-tuned models

| Label | Maronikolakis et al | | | Our results | | |
|---|---|---|---|---|---|---|
| | SVM | langBERT | mBERT | Llama 3.2 3B | Llama 3.1 8B | gpt-4o-mini |
| Derogatory | 66.8 | 85.6 | 93.1 | 72.14 | 74.85 | 75.65 |
| Exclusionary | 34.6 | 6.6 | 4.1 | 54.84 | 52.37 | 56.19 |
| Dangerous | 70.3 | 74.4 | 73.6 | 78.78 | 81.33 | 83.13 |
| F1-macro | 57.23 | 55.5 | 56.93 | 68.59 | 69.52 | 71.65 |

## 6 Conclusion

We investigated the use of LLMs for classifying extreme speech in the Indian dataset from [3]. Fine-tuning even smaller Llama models yielded significant performance improvements over SVM, langBERT, and mBERT, the best-performing methods in Maronikolakis et al. [3].

Our findings highlight the potential of open-source models to match closed-source performance while being cost-effective, reflecting rapid advancements in the open-source LLM space. Due to computational constraints, we did not explore larger mod-els like Llama 3.1 405B or GPT-4o. Future work could investigate the impact of larger models and advanced techniques like ORPO [16] for further gains.